\documentclass[runningheads]{llncs}
\usepackage{graphicx}
\usepackage[pagebackref,breaklinks,colorlinks]{hyperref}

\usepackage{amsmath,amssymb} 
\usepackage{color}

\usepackage{graphicx}
\usepackage{booktabs}

\usepackage{algorithm}
\usepackage{algpseudocode}
\usepackage{cite}
\usepackage{tikz}
\usepackage{comment}
\usepackage{multirow}

\newcommand{\etal}{\textit{et al.}}
\usepackage{soul}

\usepackage[capitalize]{cleveref}
\crefname{section}{Section}{Sections}
\Crefname{section}{Section}{Sections}
\Crefname{table}{Table}{Tables}
\crefname{table}{Tabla}{Tables}

\begin{document}
\title{Decanus to Legatus: Synthetic training for 2D-3D human pose lifting\thanks{ This work was granted access to the HPC resources of IDRIS under the allocation 2021-AD011012640 made by GENCI, and was supported and funded by Ergonova.}}
\titlerunning{Synthetic training for 2D-3D human pose lifting}
%
\author{Yue Zhu\inst{1}\orcidID{0000-0002-0914-4815} \and
David Picard\inst{1}\orcidID{0000-0002-6296-4222} }
\authorrunning{Y. Zhu, D. Picard}
%
\institute{LIGM, Ecole des Ponts, Univ Gustave Eiffel, CNRS, Marne-la-Vallée, France\\
\email{\{yue.zhu, david.picard\}@enpc.fr}\\
\url{https://github.com/Zhuyue0324/Decanus-to-Legatus}}
\maketitle 
\begin{abstract}
3D human pose estimation is a challenging task because of the difficulty to acquire ground-truth data outside of controlled environments. A number of further issues have been hindering progress in building a universal and robust model for this task, including domain gaps between different datasets, unseen actions between train and test datasets, various hardware settings and high cost of annotation, etc. In this paper, we propose an algorithm to generate infinite 3D synthetic human poses (Legatus) from a 3D pose distribution based on 10 initial handcrafted 3D poses (Decanus) during the training of a 2D to 3D human pose lifter neural network. Our results show that we can achieve 3D pose estimation performance comparable to methods using real data from specialized datasets but in a zero-shot setup, showing the generalization potential of our framework.

\keywords{3D Human pose \and Synthetic training \and Zero-shot}
\end{abstract}

\begin{figure}[t]
  \centering
  \includegraphics[width=0.8\linewidth]{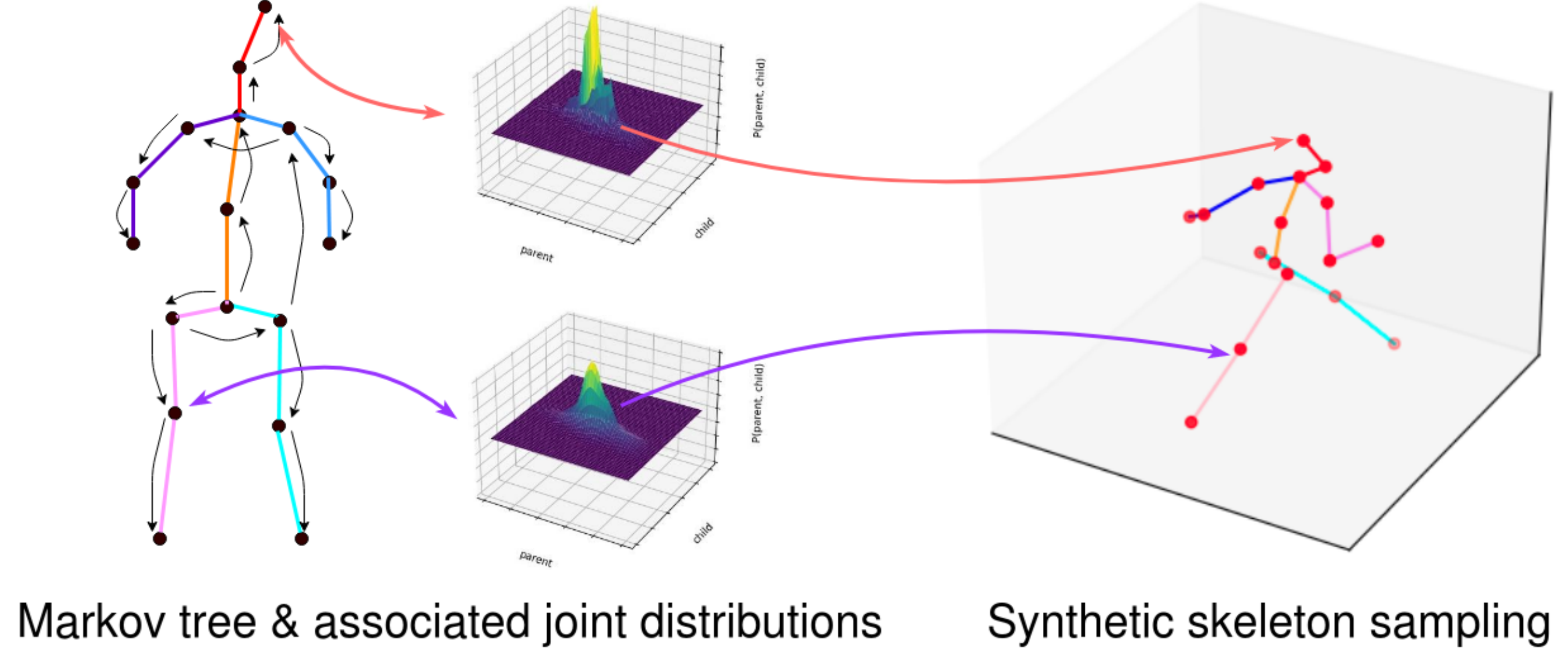}
  \caption{The main idea of our synthetic generation method: use a hierarchic probabilistic tree and its per joint distribution to generate realistic synthetic 3D human poses.}
  \label{fig:general-idea}
\end{figure}

\section{Introduction}
\label{sec:intro}

3D Human pose estimation from single images \cite{1542030(Monocular3D)} is a challenging and yet very important topic in computer vision because of its numerous applications from pedestrian movement prediction to sports analysis. Given an RGB image, the system predicts the 3D positions of the key body joints of human(s) in the image. Recent works on deep learning methods have shown very promising results on this topic\cite{DBLP:journals/corr/abs-2003-10350,Xu_2021_CVPR(GraphHourglass),DBLP:journals/corr/abs-2006-07778(cascade),Ma_2021_CVPR(ContextModeling),Xu_2020_CVPR(Kinematic), DBLP:journals/corr/abs-2007-11432}.
Current existing discriminative 3D human pose estimation methods, in which the neural network directly outputs the positions, can be put into two categories: One stage methods which directly estimate the 3D poses inside the world or camera space \cite{mehta2017monocular(3DHP), DBLP:journals/corr/PavlakosZDD16(CoarsetoFine)}, or two stage methods which first estimate 2D human poses in the camera space, then lift 2D estimated skeletons to 3D\cite{Iqbal_2020_CVPR(MultiviewInTheWild)}. 

However, all these approaches require massive amount of supervision data to train the neural network. Contrarily to 2D annotations, obtaining the 3D annotations for training and evaluating these methods is usually limited to controlled environments for technical reasons (Motion capture systems, camera calibration, etc). This brings a weakness in generalization to in-the-wild images, where there can be more unseen scenarios with different kinds of human appearances, backgrounds and camera parameters. 

In comparison, obtaining 2D annotations is much easier, and there are much more diverse existing 2D datasets in the wild\cite{DBLP:journals/corr/LinMBHPRDZ14(COCO),inproceedings(MPII),DBLP:journals/corr/abs-1803-10683(OCHuman)}. This makes 2D to 3D pose lifting very appealing since they can benefit from the more diverse 2D data at least for their 2D detection part. Since the lifting part does not require the input image but only the 2D keypoints, we infer that it can be trained without any real ground-truth 3D information.
Training 3D lifting without using explicit 3D ground-truth has previously been realized by using multiple views and cross-view consistency to ensure correct 3D reconstructions \cite{DBLP:journals/corr/abs-2011-14679(canonpose)}. However, multiple views can be cumbersome to acquire and are also limited to controlled environments.

In order to tackle this problem, we propose an algorithm which generates infinite synthetic 3D human skeletons on the fly during the training of the lifter from just a few initial handcrafted poses. This generator provides enough data to train a lifter to invert 2D projections of these generated skeletons back to 3D, and can also be used to generate multiple views for cross-view consistency. We introduce a Markov chain with a tree structure (Markov tree) type of model, following a hierarchical parent-child joint order which allows us to generate skeletons with a distribution that we evolve through time so as to increase the complexity of the generated poses (see \autoref{fig:general-idea}). We evaluate our approach on the two benchmark datasets Human3.6M and MPI-INF-3DHP and achieve zero-shot results that are competitive with that of weakly supervised methods.
To summarize, our contributions are:
\begin{itemize}
    \itemsep0em
    \item A 3D human pose generation algorithm following a probabilistic hierarchical architecture and a set of distributions, which uses zero real 3D pose data.
    \item A Markov tree model of distributions that evolve through time, allowing generation of unseen human poses.
    \item A semi-automatic way to handcraft few 3D poses to seed initial distribution.
    \item Zero-shot results that are competitive with methods using real data.
\end{itemize}

\section{Related work}
\label{sec:relatedwork}

\paragraph{Monocular 3D human pose estimation.}

In recent years, monocular 3D human pose estimation  has been widely explored in the community. The models can be mainly categorized into generative models \cite{SMPL:2015(SMPLModel),DBLP:journals/corr/BogoKLG0B16(SMPLifyModel),DBLP:journals/corr/abs-1904-05866(SMPLfromImageModel),Anguelov2005SCAPESC(SCAPEModel),Akhter_2015_CVPR(PoseConditioned),Schmidtke_2021_CVPR(EclipseBodyJoint),9157563} which fit 3D parametric models to the image, and discriminative models which directly learn 3D positions from image\cite{1542030(Monocular3D),DBLP:journals/corr/abs-1803-04775(LearnMultiview)}. Generative models try to fit the shape of the entire body and as such are great for augmented reality or animation purpose\cite{petrovich21actor(Mathis)}. However, they tend to be less precise than discriminative models.
On the other hand, a difficulty that the discriminative models have is that depth information is hard to infer from a single image when it is not explicitly modeled, and thus additional bias must be learned using 3D supervision \cite{Luo_2021_CVPR(InterCarpet), Ma_2021_CVPR(ContextModeling)}, multiview spatial consistency \cite{Fang_2021_CVPR(Mirror),Xu_2020_CVPR(Kinematic), DBLP:journals/corr/abs-2011-14679(canonpose)} or temporal consistency\cite{1542030(Monocular3D),Liu_2020_CVPR(PoseReconstruction), Choi_2021_CVPR(TempConsistency}.
Discriminative models can also be categorized into one stage models which predict directly 3D poses from images\cite{mehta2017monocular(3DHP),DBLP:journals/corr/PavlakosZDD16(CoarsetoFine),Luo_2021_CVPR(InterCarpet), DBLP:journals/corr/abs-2001-02024} and two stage methods which first learn a 2D pose estimator, then lift the obtained 2D poses to 3D \cite{DBLP:journals/corr/ZhouZLDD15,DBLP:journals/corr/MartinezHRL17, Xu_2020_CVPR(Kinematic), Iqbal_2020_CVPR(MultiviewInTheWild), Xu_2021_CVPR(GraphHourglass),DBLP:journals/corr/abs-2011-14679(canonpose)}. 
Lifting 2D pose to 3D is somewhat of an ill-posed problem because of depth ambiguity ambiguity. But the larger quantity and diversity of 2D datasets\cite{DBLP:journals/corr/LinMBHPRDZ14(COCO), inproceedings(MPII), DBLP:journals/corr/abs-1803-10683(OCHuman)}, as well as the already achieved much better performance in 2D human pose estimation provide a strong argument for focusing on lifting 2D human poses to 3D. 

\paragraph{Weak supervision methods.}

Since obtaining precise 3D annotations of human poses are hard due to technical reasons and are mostly limited to controlled environments, many research proposals tackled this problem by designing weak supervision methods to avoid using 3D annotations. For example, Iqbal et al. \cite{Iqbal_2020_CVPR(MultiviewInTheWild)} apply a rigid-aligned multiview consistency 3D loss between multiple 3D poses estimated from different 2D views of the same 3D sample. Mitra et al. \cite{Mitra_2020_CVPR(AnchorMetric)} learn 3D pose in a canonical form and ensure same predicted poses from different views. 
Fang et al. \cite{Fang_2021_CVPR(Mirror)} propose a virtual mirror so that the estimated 3D poses, after being symmetrically projected into the other side of the mirror, should also look correctly, thus simulating another way of ‘multiview’ consistency. Finally, Wandt et al. \cite{DBLP:journals/corr/abs-2011-14679(canonpose)} learn lifted 3D poses in a canonical form as well as a camera position so that every 3D pose lifted from a different view of a same 3D sample should still have 2D reprojection consistencies. For us,  in addition to 3D supervision obtained from our synthetical generation, we also use multiview consistency to improve our training performance.

\paragraph{Synthetic human pose training.} 

Since the early days of the Kinect, synthetic training has been a popular option for estimating 3D human body pose 
\cite{shotton2011real(realtimeposerecog)}.
The most common strategy is to perform data augmentation in order to increase the 
size and diversity of real  datasets\cite{Gong_2021_CVPR(PoseAug)}. Others like
Sminchisescu \etal \cite{1640965} render synthetically generated poses on natural indoor and outdoor image backgrounds.
Okada \etal \cite{Okada2008RelevantFS} generate synthetic human poses in a subspace constructed by PCA using the walking sequences extracted from the CMU Mocap dataset\cite{cmu-mocap-dataset}.
Du \etal \cite{Du2016MarkerLess3H} create a synthetic height-map dataset to train a dual-stream convolutional network for 2D joints localization.
Ghezelghieh \etal \cite{DBLP:journals/corr/GhezelghiehKS16} utilize 3D graphic software and the CMU Mocap dataset to synthesize humans with different 3D poses and viewpoints.
Pumarola \etal \cite{DBLP:journals/corr/abs-1904-04571(3DPeopleDataset)} created 3DPeople, a large-scale synthetic dataset of photo-realistic images with a large variety of subjects, activities and human outfits.
Both \cite{Clever_2020_CVPR(BodyAtRest)} and \cite{Luo_2021_CVPR(InterCarpet)} use pressure maps as input to estimate 3D human pose with synthetic data. 
In this paper, we are only interested in generating realistic 3D poses as a set of keypoints so as to train a 2D to 3D lifting neural network. As such, we do not need to render visually realistic humans with meshes, textures and colors for this much simpler task.

\paragraph{Human pose prior.}

Since the human body is highly constrained, it can be leveraged as an inductive bias in pose estimation.
Bregler\etal\cite{Bregler1998TrackingPW} use kinematic-chain human pose model that follow the skeletal structure, extended by Sigal \etal\cite{Sigal:IJCV:11} with interpenetration constraints.
Chow\etal\cite{Chow-Liu-Tree} introduced Chow-Liu tree, the maximum spanning tree of all-pairwise-mutual-information tree to model pairs of joints that exhibit a high flow of information.
Lehrmann\etal\cite{Lehrmann2013Non-parametric} use a Chow-Liu tree that maximize an entropy function depending on nearest neighbor distances and learn local  conditional  distributions from data based on this tree structure.
Sidenblahn\etal\cite{Black:ECCV:2000} use cylinders and spheres to model human body.
Akhter \etal \cite{Akhter_2015_CVPR(PoseConditioned)} learn joint-angle limits prior under local coordinate systems of 3 human body parts as torso, head,and upper-legs. We use a variant of kinematic model because the 3D limb lengths are fixed no matter the view, which can facilitate the generation process of synthetic skeleton.

\paragraph{Cross dataset generalization.}
Due to the diversity of human appearances and view points, cross-dataset generalization has recently been the center of attention of several works.
Wang \etal \cite{DBLP:journals/corr/abs-2004-03143(predict-camera-view-CD)} learn to predict camera views so as to auto-adjust to different datasets.  
Li \etal\cite{DBLP:journals/corr/abs-2006-07778(cascade)} and Gong \etal \cite{Gong_2021_CVPR(PoseAug)} perform data augmentation to cover the possible unseen poses in test dataset.
Rapczyński \etal \cite{s21113769} discuss several methods including normalisation, viewpoint estimation, etc., for improving cross-dataset generalization. 
In our method, since we use purely synthetic data, we are always in a cross-dataset generalization setup.

\section{Proposed method}
\label{sec:ownwork}

The goal of our method is to create a simple synthetic human pose generation model allowing us to train on pure synthetic data without any real 3D human pose data information during the whole training procedure.

\subsection{Synthetic human pose generation model}
\label{sec:ownwork-generator}

\subsubsection{Local spherical coordinate system.}

\begin{figure}[tb]
  \centering
  \includegraphics[width=0.8\linewidth]{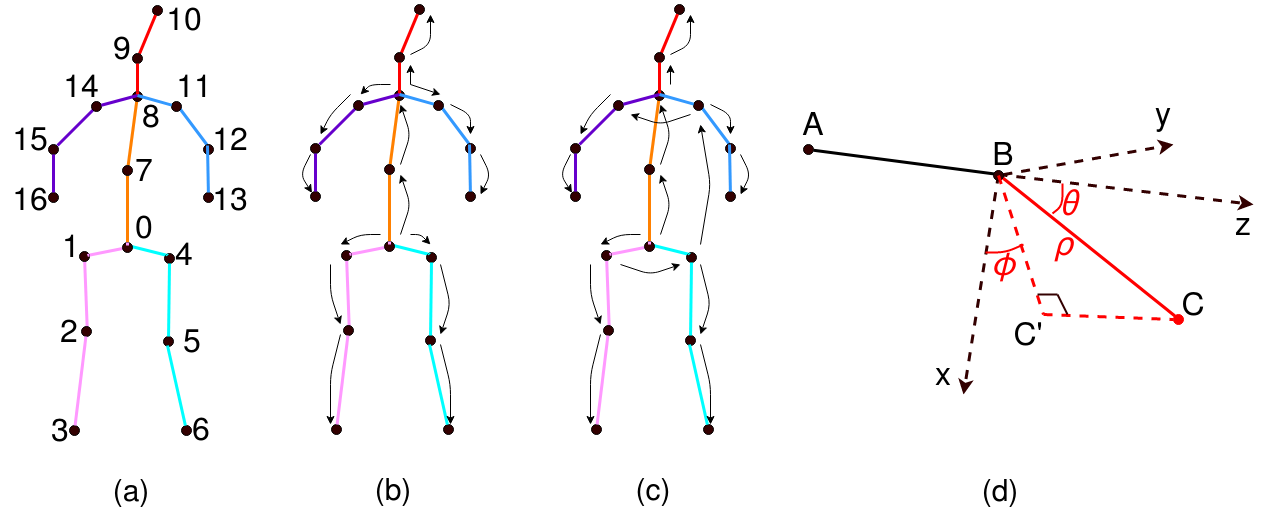}
  \caption{(a) The 17-joint model of Human3.6M that we use (b) The \textbf{parent-child joint relation} graph. With parent joint's coordinate as origin of local spherical coordinate system, it generates child joint's position. (c) The \textbf{parent-child $\rho$, $\theta$ , $\phi$ relation} graph. With parent joint's $\rho$, $\theta$ , $\phi$ information, it samples child joint's $\rho$, $\theta$ , $\phi$. (d) An example of how child joint is generated with sampled $\rho$, $\theta$ , $\phi$ from relationship in (c) under the local spherical coordinate system with it's parent joint in (b) as origin.}
  \label{fig:skeleton-generation}
\end{figure}

Without loss of generalization, we use Human3.6M skeleton layout  shown in \autoref{fig:skeleton-generation} (a) throughout the paper.
To simplify human pose generation, we set the pelvis joint (joint 0) as root joint and the origin of the global Cartesian coordinate system from which a tree structure is applied to generate joints one by one. We suppose that the position of one joint depends on the position of the joint which is directly connected to it but closer (in geodesic meaning) to the root joint. We call this kinematic chain \textbf{parent-child joint relations}, as shown in \autoref{fig:skeleton-generation} (b). With this relationship, we propose to generate the child joint in a local spherical coordinate system ($\rho, \theta, \phi$) centered on its parent joint (see \autoref{fig:skeleton-generation} (d)). The $\rho$, $\theta$ , $\phi$ values are sampled with respect to a conditional distribution $P(x_{child}|x_{parent})$. This produces a Markov chain indexed by a tree structure which we denote as a Markov Tree. 

Our motivation to use a local spherical coordinate system for joint generation is that each human body branch has a fixed length $\rho$ no matter the movement. Also, since the supination and the pronation of the branches are not encoded in skeleton representation, the new joint position can be parameterized with polar angle $\theta$ and azimuthal angle $\phi$. Furthermore, by using an axis system depending on 'grandparent-parent' branch instead of global coordinate system, the possible angle interval of $\theta$ and $\phi$ achieved by human is more limited than in a global coordinate system. Finally, our local spherical coordinate system is entirely bijective with global coordinate system. 

\subsubsection{Hierarchic probabilistic skeleton sampling model.}
\label{subsec:distribution}

Generating a human pose in our local spherical coordinate system is equivalent to generating a set of ($\rho$,$\theta$,$\phi$). We thus propose to sample these values from a distribution that approximate that of real human poses. To retain plausible poses, we limit the range of ($\rho$,$\theta$,$\phi$) for each joint based on what is on average biologically achievable.

Since body joints follow a tree-like structure, it is unlikely that sampling each joint independently of the others leads to realistic poses. Instead, we propose to model the distribution of the joints by a Markov chain index by a tree following the skeleton, where probability of sampling a tuple ($\rho$,$\theta$,$\phi$) for a joint depends on the values sampled for its parent. More formally, denoting a child joint $c$ and its parent $p(c)$ following the tree structure, we have:
\begin{equation}
(\rho_c, \theta_c, \phi_c) \sim P((\rho, \theta, \phi) | (\rho_{p(c)}, \theta_{p(c)}, \phi_{p(c)}))
\end{equation}
Please note that the tree structure used for accounting the dependencies between joints as shown on \autoref{fig:skeleton-generation} (c) is slightly different than the kinematic one. We found in practice that it is better to condition the position of one shoulder on the position of the same side hip, and to condition symmetrical shoulder/hip on their already generated counterpart rather than on their common parent. Intuitively, this seems to better encode global consistency.

To facilitate modeling distribution $P((\rho,\theta,\phi)|(\rho_{p(c)},\theta_{p(c)},\phi_{p(c)}))$, we make further assumption that all 3 components only depend on their parent counterparts. More formally:

\begin{align}
\rho_c \sim P(\rho|\rho_{p(c)}),\  
\theta_c \sim P(\theta|\theta_{p(c)}),\ 
\phi_c \sim P(\phi|\phi_{p(c)})
\end{align}

This allows us to model each distribution with a simple non-parametric model consisting of a simple 2D histogram representing the probability of sampling, \textit{e.g.}, $\rho_c$ knowing the value of $\rho_{p(c)}$. In practice, we use 50 bins histograms for each value, totalling to $3\times 16 = 48$ 2D histograms of size $50\times 50$. When there is no ambiguity, we use the same notation $P(\cdot|\cdot)$ for the histogram and the probability.

\subsection{Pseudo-realistic 3D human pose sampling}

The next step is to estimate a distribution that can approximate the real 3D pose distribution, and from which our model can sample, so that the generated poses look like real human actions. Under the constraint of zero-shot 3D real data, we choose to make breakthrough by looking at limited amount of 2D real poses and 'manually' lift them into 3D to make our distribution. However, it is impossible for us to tell the exact depths of keypoints from an image with our eye, and it is also a huge amount of work to do if we check a lot of images one by one. Instead, we choose a 3-step procedure to get our handcrafted 3D pose: 

\subsubsection{High-variance 2D poses.}\label{subsec:initialpose}
We randomly sample 1000 sets of 10 2D-human poses from the target dataset (\textit{e.g.}, Human3.6M). We then compute the total variance for each set and pick the sets with largest variance as our candidates. This ensure our initial pose set has high diversity.

\subsubsection{Semi-automatic 2D to 3D seed pose lifting.}
Next, we use a semi-automatic way to lift samples in each seed set to 3D.
The idea is as follows: from an image for which we already know the 2D distances between connected joints, and if we can estimate the 3D length of each branch who connects the joints as well as the proportion $\lambda_{prop}$ between the 2D length in the image (in pixel) and the 3D length (in centimeter), we can estimate the relative depth between connected joints using Pythagorean theorems under the assumption that the camera produces an almost orthogonal projection. The ambiguity about the sign of these depths, which decide if one joint is in front of or in the back of its parent joint, can easily be manually annotated.

\begin{figure}[b]
  \centering
  \includegraphics[width=0.7\linewidth]{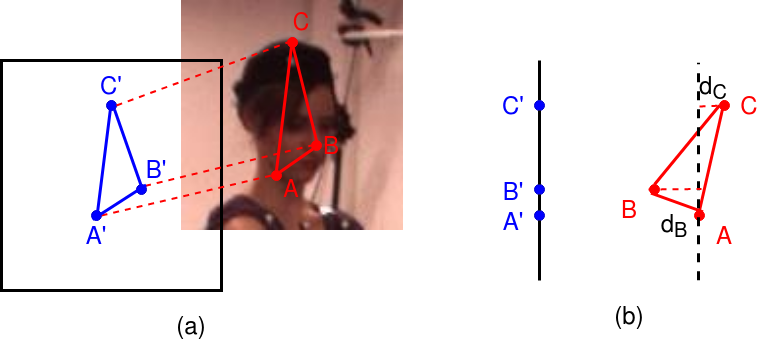}
  \caption{\textbf{(a)} 3D poses (red $A$,$B$ and $C$, unit in centimeters) of 3 joints of the head projected onto 2D camera plan (blue $A'$,$B'$ and $C'$, unit in pixels). \textbf{(b)} same but right side view after $90^{o}$ rotation.}
  \label{fig:headtriangle}
\end{figure}

To estimate the 3D length, we define a set of fixed value representing branch lengths ($||c-p(c)||_2, \forall c$ except the root joint) of the human body based on biological data. 
Since we later calculate under a proportionality assumption between 3D and 2D, we only need it to roughly represent the proportionality between different human bone length. We also manually annotate $sign_c$ for each keypoint $c$, denoting if it is relatively further or closer to the camera compared to its parent joint $p(c)$. 
Finally the 2D-3D size proportion $\lambda_{prop}$ is calculated under the assumption that the 3 joints around the head (head top, nose and neck) form a triangle of known ratio which is independent of rotation and view, visually shown in \autoref{fig:headtriangle}. This is reasonable since there are no largely moving articulated part in this triplet. We choose $AB=1$ the unit length and we suppose the proportion between $AB$, $BC$ and $CA$ is fixed ($BC=\alpha AB$,$AC=\beta AB$). Noting $d_B=B'B-A'A$ and $d_C=C'C-A'A$, for the 2D skeleton we know $A'B'$,$B'C'$ and $A'C'$, then we have 3 unknown variables $d_B$, $d_C$, and $\lambda_{prop}=\frac{A'B'(pixels)}{A'B'(meters)}$ and 3 equations: 
\begin{align}
\nonumber d_B^2=AB^2-(\frac{A'B'}{\lambda_{prop}})^2&,\quad d_C^2=(\beta AB)^2-(\frac{A'C'}{\lambda_{prop}})^2,\\
(d_B-d_C)^2&=(\alpha AB)^2-(\frac{B'C'}{\lambda_{prop}})^2
\end{align} 
Then we can solve $\lambda_{prop}$. In practice, we set $\alpha=1$ and $\beta=5/3$.

\begin{figure*}[tb]
  \centering
  \includegraphics[width=0.8\linewidth]{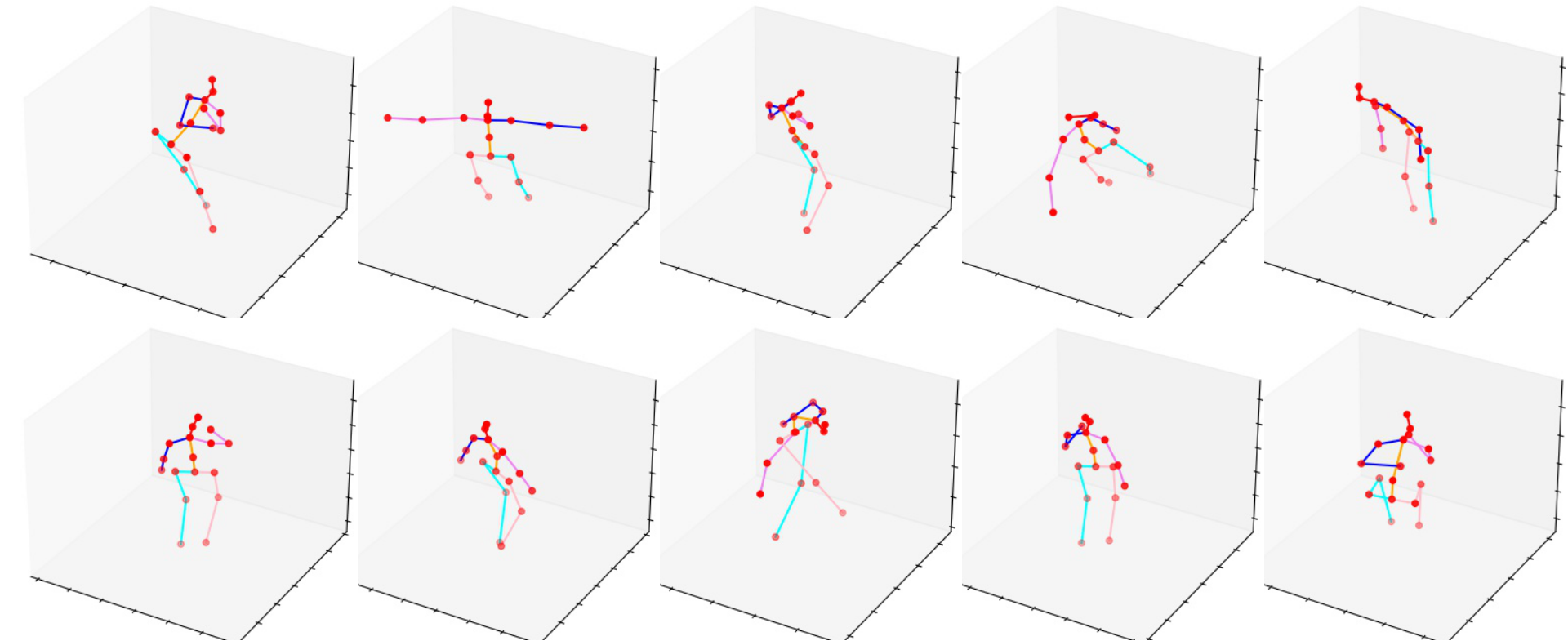}
  \caption{A example of a set of 10 semi-automatic lifted 3D poses.
  This set of seeds is also the one which produce our best score on Human3.6M dataset. These 10 lifted samples have a 79.42mm MPJPE error compare to the groundtruth.}
  \label{fig:init10poses}
\end{figure*}

After obtaining these depths, we apply Pythagorean theorem to get the final depth value of all joints with the kinematic order. Examples of semi-automatic lifted 3D poses are shown on \autoref{fig:init10poses}. 
Since there are only a few keypoints to label as \emph{in front of} or \emph{behind} their parent joint, the labeling process is very easy and takes about 3 minutes per image only.

\subsubsection{Distribution diffusion}

We then transform 3D poses into the local spherical coordinate system and used each seed set as initial distribution to populate the histograms. Since the sampling of a new skeleton follows the Markov tree structure and different limbs have a weak correlation between them in our model, it is possible to sample skeletons that look like combinations of the original 10 samples within the seed set.

However, these initial samplings are by no mean complete, and we run the risk of overfitting the lifter network to these poses only. To alleviate this problem, we introduce a diffusion process among each 2D histogram such that the probability of adjacent parameters is raised over time. More formally:

\begin{align}
  P(x_c|x_{p(c)})_{t+1} = P(x_c|x_{p(c)})_t + \alpha_{x_c} \Delta P(x_c|x_{p(c)})_t,\ x\in\{\rho, \theta, \phi\}
\label{eq:diffusion}
\end{align}
where $\Delta$ is the Laplacien operator and $\alpha_{x_c}$ is the diffusion coefficient. This idea is derived from the heat diffusion equation in thermodynamics, in which bins with a higher probability diffuse to their neighbours (Laplacian operator), making the generation process more and more likely to generate samples out of initial bin.  

\begin{figure}[tb]
  \centering
  \includegraphics[width=0.9\linewidth]{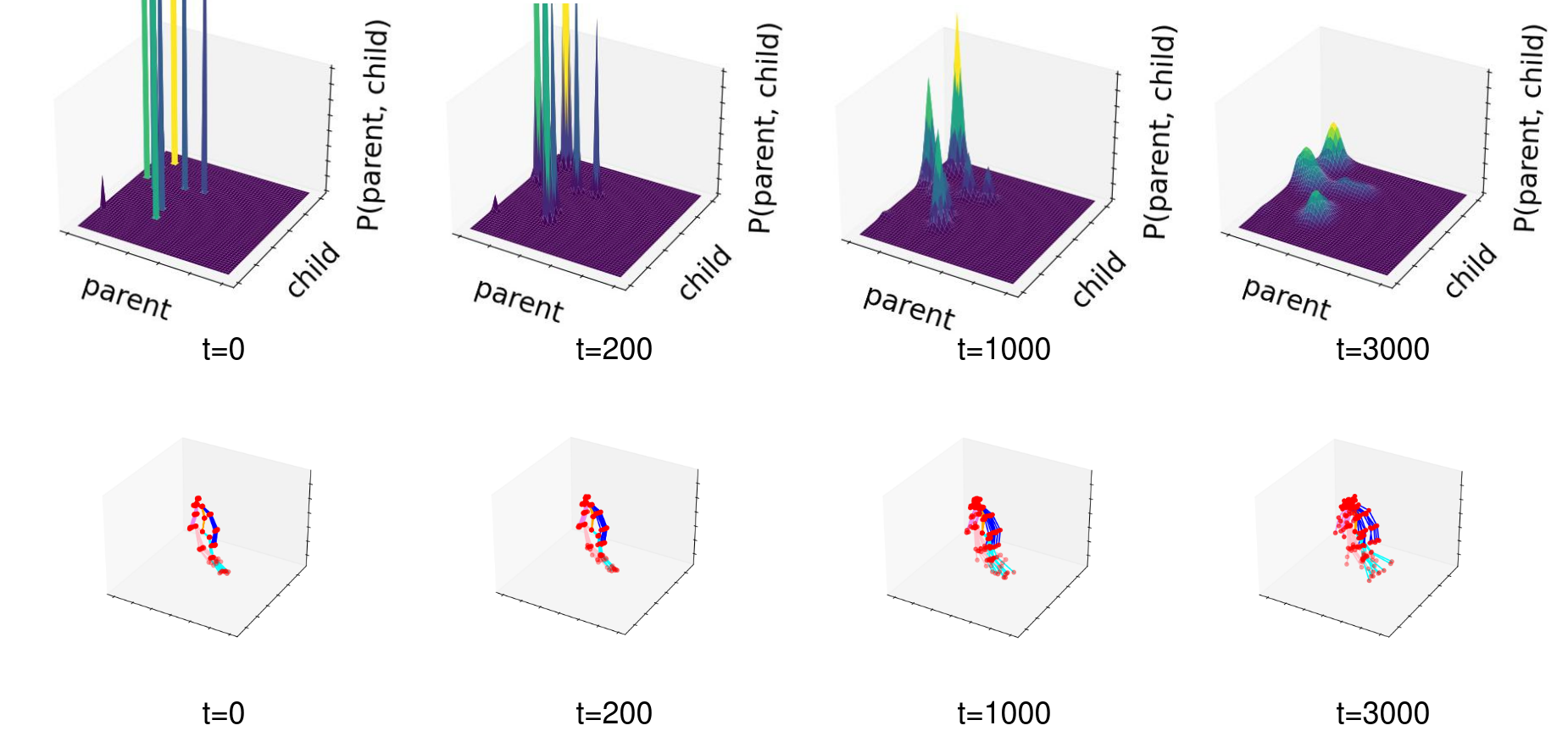}
  
  \caption{First row is an example of the distribution histogram of a joint after 0, 200, 1000 and 3000 steps of diffusion. Second row shows an example of slightly increased generation variety when sampling from a single bin and generating 10 samples each time after 0, 200, 1000 and 3000 steps of diffusion.
  }
  \label{fig:difusionsample}
\end{figure}

The main reason behind our diffusion process is that of curriculum learning\cite{bengio2009curriculum}. At first, the diversity of sampled skeletons is low and the neural network is able to quickly learn how to lift these poses. At later stage, the diffusion process allows the sampling process to generate more diverse skeletons that are progressive extensions of the initial pose angles, avoiding overfitting the original poses. We show in \autoref{fig:difusionsample} an example of evolution of the histogram and increase of generation variety through diffusion.

\subsection{Training with synthetic data}
\label{sec:trainsynt}

\begin{figure*}[tb]
  \centering
  \includegraphics[width=\linewidth]{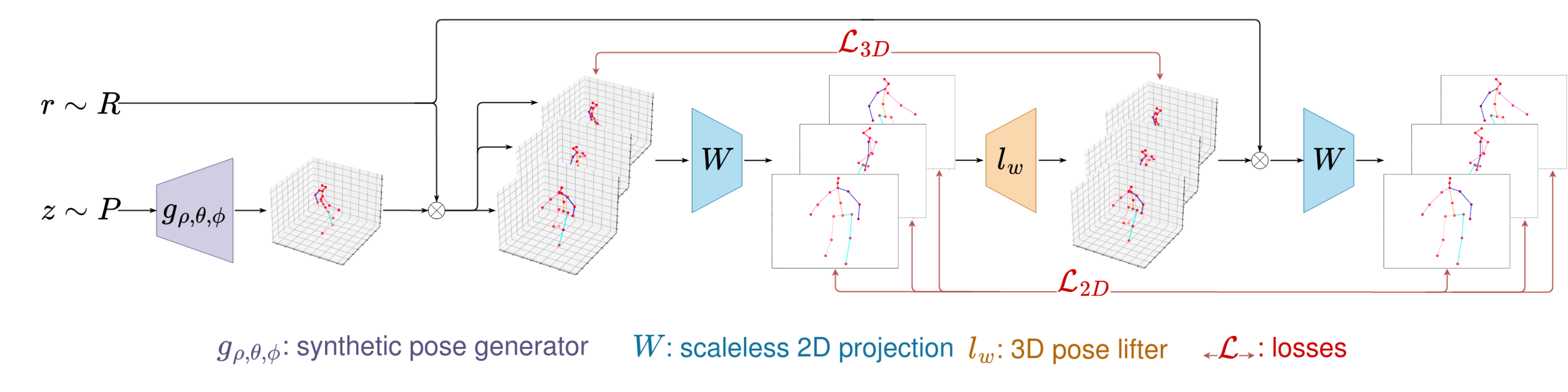}
  \caption{Our whole training process with synthetic data. Our generator $g$ generates a 3D human pose following given distributions $P$ of $\rho$, $\theta$ and $\phi$. It will be applied with multiple different random generated $r$ to project into different camera view. Projector $W$ will projects them into scaleless 2D coordinates and they are the network inputs. The output estimated 3D poses will be applied with scaleless 3D supervision loss $\mathcal{L}_{3D}$, and also cross-view scaleless 2D reprojection loss $\mathcal{L}_{2D}$, which rotate estimated 3D pose from one view to another with known $r$ and apply 2D supervision after projection $W$.
  }
  \label{fig:trainingprocess}
\end{figure*}

The training setup of 2D-3D lifter network $l_w$ is shown on \autoref{fig:trainingprocess} and consists of 3 main components: (1) Sampling a batch of skeletons at each step ; (2) sampling different virtual cameras to project the generated skeletons into 2D ; and finally (3) the different losses used to optimize $l_w$. In practice, $l_w$ is a simple 8-layer MLP with 1 in-layer, 3 basic residual blocks of width 1024, and 1 out-layer, adapted from \cite{DBLP:journals/corr/abs-2011-14679(canonpose)}.

When sampling a new batch of skeleton using our generator, we have to keep in mind that the distribution of the generator varies through time because of the diffusion process introduced in \autoref{eq:diffusion}. To avoid over-sampling or under-sampling bins with low density, we propose to track the amount of skeletons that have been generated in each bin and adjust the sampling strategy accordingly.
More formally, let us denote $P_t$ the \emph{true distribution} obtained by \autoref{eq:diffusion}, and $P_e$ the \emph{empirical distribution} obtained by tracking the generation process. The corrected sampling algorithm is shown in \autoref{alg:sampling} and basically selects uniformly a plausible bin ($P_t > 0$) that has not been over-sampled ($P_e \leq P_t$). The whole generation process simply loops over all joints using the Markov tree and is shown on \autoref{alg:generation}.

\begin{algorithm}[t]
\caption{Sampling algorithm}\label{alg:sampling}
\begin{algorithmic}
\Require True distribution $P_t$, empirical distribution $P_e$;
\State $bins\gets$ where $P_t > 0$ and $P_e \leq P_t$
\State $b \sim \mathcal{U}(bins)$
\State \Return Random sample from $b$
\end{algorithmic}
\end{algorithm}

\begin{algorithm}
\caption{Pose generation algorithm}\label{alg:generation}
\begin{algorithmic}
\Require True distribution $P_t$, empirical distribution $P_e$, 
Markov tree structure $T$, sampling algorithm S
\State $X \gets 0_{(J,3)}$ \Comment{3=$\rho,\theta,\phi$}

\For{$i \in \rho,\theta,\phi$} \Comment{root joint}
\State $X[0,i] \gets$ S( $P_t(X_0), P_e(X_0)$)
\EndFor

\For{(p,c) in $T$} \Comment{parent-child relations in $T$}
\For{$i \in \rho,\theta,\phi$}

\State $X[c,i] \gets$ S($P_t(X_{(c,i)}|X_{(p,i)}),P_e(X_{(c,i)}|X_{(p,i)})$)
\State Update $P_e(X_{(c,i)}|X_{(p,i)})$
\EndFor
\EndFor
\State \Return $X$ in Cartesian coordinates
\end{algorithmic}
\end{algorithm}

At initialization, we sample 5000 real 2D poses, compute the proportion of nearest neighbour within each pose seed, and use it to initialize the histogram to give more importance to more frequent poses.

Regarding the projection of the batch into 2D, we propose to sample a set of batch-wise rotation matrices $R_{1, \ldots, N}$, mostly rotating around the vertical axis, to simulate different viewpoints. Then, the rotated 3D skeletons are just simply:
    $X_{3D, i} = R_iX_{3D,0}, \quad i\in\{1, \ldots, N\}$, 
with $X_{3D, 0}$ being the original skeleton in global Cartesian coordinates. To simulate the cameras, we follow \cite{DBLP:journals/corr/abs-2011-14679(canonpose)} and use a scaleless orthogonal projection:
\begin{align}
    X_{2D,i} = \frac{WX_{3D,i}}{\|WX_{3D,i}\|_F}, \quad W = \begin{pmatrix}
1 & 0 & 0\\
0 & 1 & 0
\end{pmatrix},
\label{eq:2Dproj}
\end{align}
where $W$ is the orthogonal projection matrix and $\|\cdot\|_F$ is the Frobenius norm.
Normalizing by the Frobenius norm allows us to be independent of the global scale of $X_{2D,i}$ while retaining the relative scale of each bone with respect to each other.
In practice, we found that uniformly sampling random rotation matrices at each batch renders the training much more difficult. Instead, we sample view with a small noise around the identity matrix and let the noise increase as the training goes on to generate more complex views at later stages.

Finally, to train the network, we leverage several losses. First, since we have the 3D ground-truth associated with each generated skeleton:
\begin{align}
    \mathcal{L}_{3D} =  \frac{1}{N}\underset{i=1..N}{\sum}\left\|\frac{\hat{X}_{3D,i}}{\|\hat{X}_{3D,i}\|_{F}}-\frac{X_{3D,i}}{\|X_{3D,i}\|_{F}}\right\|_1,
  \label{eq:loss3Dsynt}
\end{align}
with $\hat{X}_{3D,i} = l_w(X_{2D,i})$ being the output of the lifter $l_w$, and $\|\cdot\|_1$ the $\ell_1$ norm. 3D skeletons are normalized before being compared because the input of the lifter is scaleless and as such it would make no sense to expect the lifter to recover the global scale of $X_{3D}$.
Then, we use the multiple views generated thanks to $R_i$ to enforce a multiview consistency loss. Calling $\hat{X}_{2D,i,j}=WR_jR^{-1}_i\hat{X}_{3D,i}$ the projection of the lifted skeleton from view $i$ into view $j$, we optimize the cross-view projection error:
\begin{align}
    \mathcal{L}_{2D} = \frac{1}{N^2}\sum_{i=1}^N\sum_{j=1}^N\left\|\frac{\hat{X}_{2D,i,j}}{\|\hat{X}_{2D,i,j}\|_{F}}-\frac{X_{2D,j}}{\|X_{2D,j}\|_{F}}\right\|_1
  \label{eq:loss2Dsynt}
\end{align}
The global synthetic training loss we use is the following combination:
\begin{align}
  \mathcal{L} = \mathcal{L}_{2D}+\lambda_{3D} \mathcal{L}_{3D}
  \label{eq:losssynt}
\end{align}

\section{Experiments}
\label{sec:results}

\subsection{Datasets}
We use two widely used dataset Human3.6M\cite{journals/pami/IonescuPOS14(H36m)} and MPI-INF-3DHP\cite{mehta2017monocular(3DHP)} to quantitatively evaluate our method.

We only use our generated synthetic samples for training and evaluate on S9 and S11 of Human3.6M and TS1-TS6 on MPI-INF-3DHP with their common protocols.
In order to compare the quality of our generated skeletons with real 2D data, We also use the COCO\cite{DBLP:journals/corr/LinMBHPRDZ14(COCO)} and MPII\cite{inproceedings(MPII)} datasets to check the generalizability of our method with qualitative evaluation.

\subsection{Evaluation metrics}
For the quantitative evaluation on both Human3.6M and MPI-INF-3DHP we use MPJPE, i.e. the mean euclidean distance between the reconstructed
and ground-truth 3D pose coordinates after the root joint is aligned ($P1$ evaluation protocol of Human3.6M dataset).
Since we train the network with a scaleless loss, we follow \cite{DBLP:journals/corr/abs-2011-14679(canonpose)} and 
scale the output 3D pose's Forbenius norm into the ground-truth 3D pose's Forbenius norm in order to compute the MPJPE. 
We also report PCK, i.e. the percentage of keypoints with the distance between predicted 3D pose and ground-truth 3D pose is less or equal to half of the head's length.

\subsection{Implementation details}
We use a batch-size of 32 and we train for 10 epochs on a single 16G GPU
using Adam optimizer and a learning rate of $10^{-4}$. We set the number of views $N=4$ and the total number of synthetic 2D input samples for each epoch is the same as the number of H36M training samples to make a fair comparison.
The distribution diffusion coefficient $\alpha_{x_c}$ is a joint-wise loss dependent value, set to $10^{-5}\times 10^{|\delta \mathcal{L} |/(10\times N)}$ where $\delta \mathcal{L}$ is the joint-wise difference between loss of the last batch and the current batch, and the rotation
$R$ are sampled with a noise that increases in
$\frac{1}{2\times \# batch}$
after each step, with 
$\# batch$ the number of elapsed batches in the current epoch. 
For the loss, $\lambda_{3D}=0.1$ is set empirically.
To account for the variation due to the selection of the 2D pose using total variance, we keep the 10 sets with highest variance and show averaged results. Our method trains on about 100k generated samples per hour on a V100 GPU, whereas inference time for lifting is negligible.

\subsection{Comparison with the state-of-the art}

We compare our results with the state-of-the-art methods with synthetic supervision for training in 
\autoref{tab:sota}. We present several weak supervision methods which also do not use real 3D annotations, and instead use other sort of real data supervision whereas we do not.
We can see that our method outperforms these synthetic training methods and achieves the performance on par with weakly supervised methods on H36M, while never using a real example for training.

\begin{table}[t]
    \small
    \centering
    \begin{tabular}{c|c|ccc|cccc|cc}
    \multicolumn{2}{c}{} & \multicolumn{3}{c|}{Weak supervision} & \multicolumn{4}{c|}{Synthetic training} & \multicolumn{2}{c}{\textbf{Ours}} \\
    \cline{3-11}
 \multicolumn{2}{c|}{} &  \cite{Iqbal_2020_CVPR(MultiviewInTheWild)} &  \cite{Mitra_2020_CVPR(AnchorMetric)} &  \cite{DBLP:journals/corr/abs-2011-14679(canonpose)}  &  \cite{DBLP:journals/corr/abs-2006-07778(cascade)} &  \cite{DBLP:journals/corr/GhezelghiehKS16} &  \cite{Du2016MarkerLess3H} &  \cite{varol17_surreal(SURREAL)} &  10 sets &   best run \\
 \hline
H36M &  MPJPE$\downarrow$ &    67.4 &  120.95 &  65.9 &    106.8 &  $\geq$ 78.13 &  126.47 &  111.6 &  95.4$\pm$13.5 &  60.8 \\
\multirow{2}{*}{3DHP} &  MPJPE$\downarrow$ &    109.3 &  - &  104.0 &    - &  - &  - &  - &  148.4$\pm$7.6 &  132.8 \\
 &  PCK$\uparrow$ &    79.5 &  - &  77.0 &    - &  - &  - &  - &  57.7$\pm$2.3 &  61.9
\end{tabular}
    \caption{Comparison of our results with the state-of-the-arts under the common protocol 1 on Human 3.6M and MPI-INF-3DHP. The value before and after $\pm$ symbol are mean and standard deviation values.}
    \label{tab:sota}
\end{table}

\begin{figure*}[tb]
  \centering
  \includegraphics[width=0.9\linewidth]{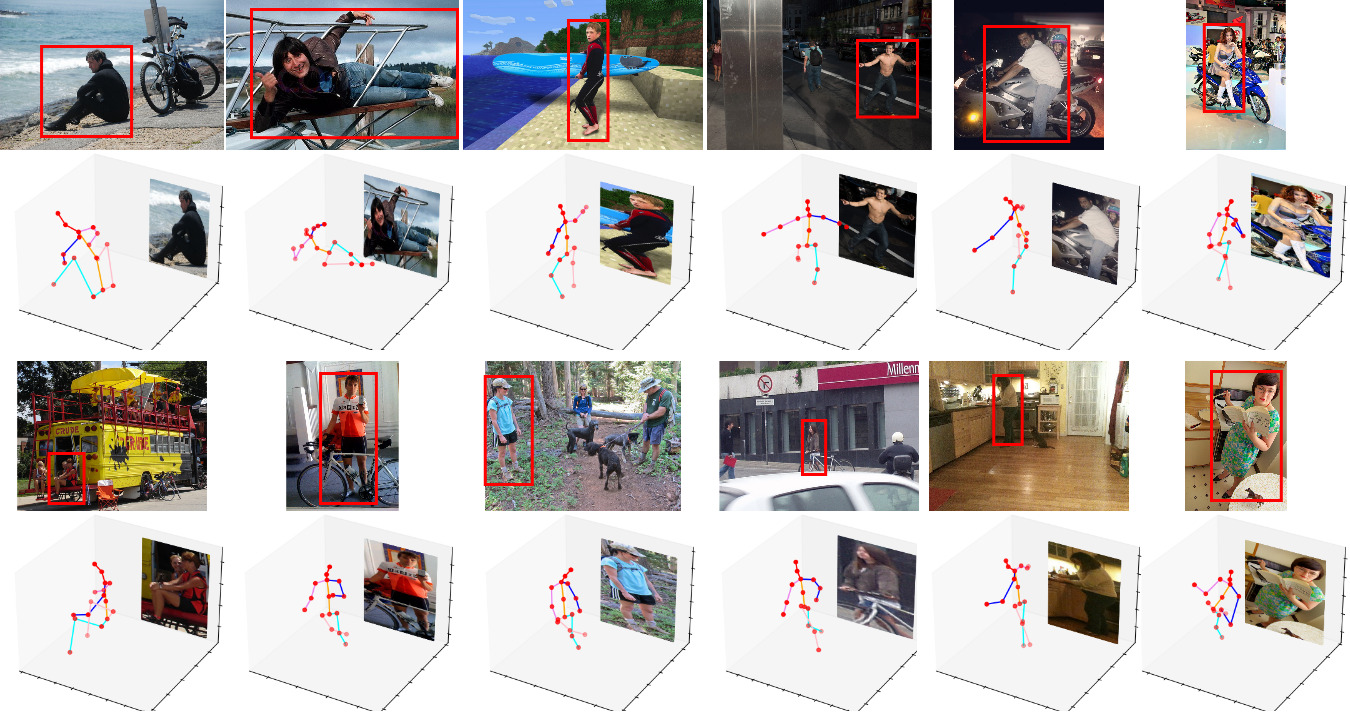}
  \caption{Example of zero shot lifting in the wild on images from the COCO dataset. The first row are visually correct prediction, while the last row presents 'failure' cases, mostly due to right leg learnt a bias of leaning backward.}
  \label{fig:qualitativeresult}
\end{figure*}

We show qualitative results on the COCO dataset on \autoref{fig:qualitativeresult}. 
Since the COCO layout is different from that of H36M, we use a linear interpolation of existing joints to localize the missing joints. We can see that our model still achieves good qualitative performances on zero shot lifting of human poses in the wild (first 2 rows). Failed predictions (last row) tend to bend the legs backward even when the human is standing still, which may be a bias of the generator.

\section{Ablation studies}

\subsection{Synthetic poses realism}\label{subsec: pre-rec-methodology}

\begin{figure}[t]
  \centering
  \includegraphics[width=0.5\linewidth]{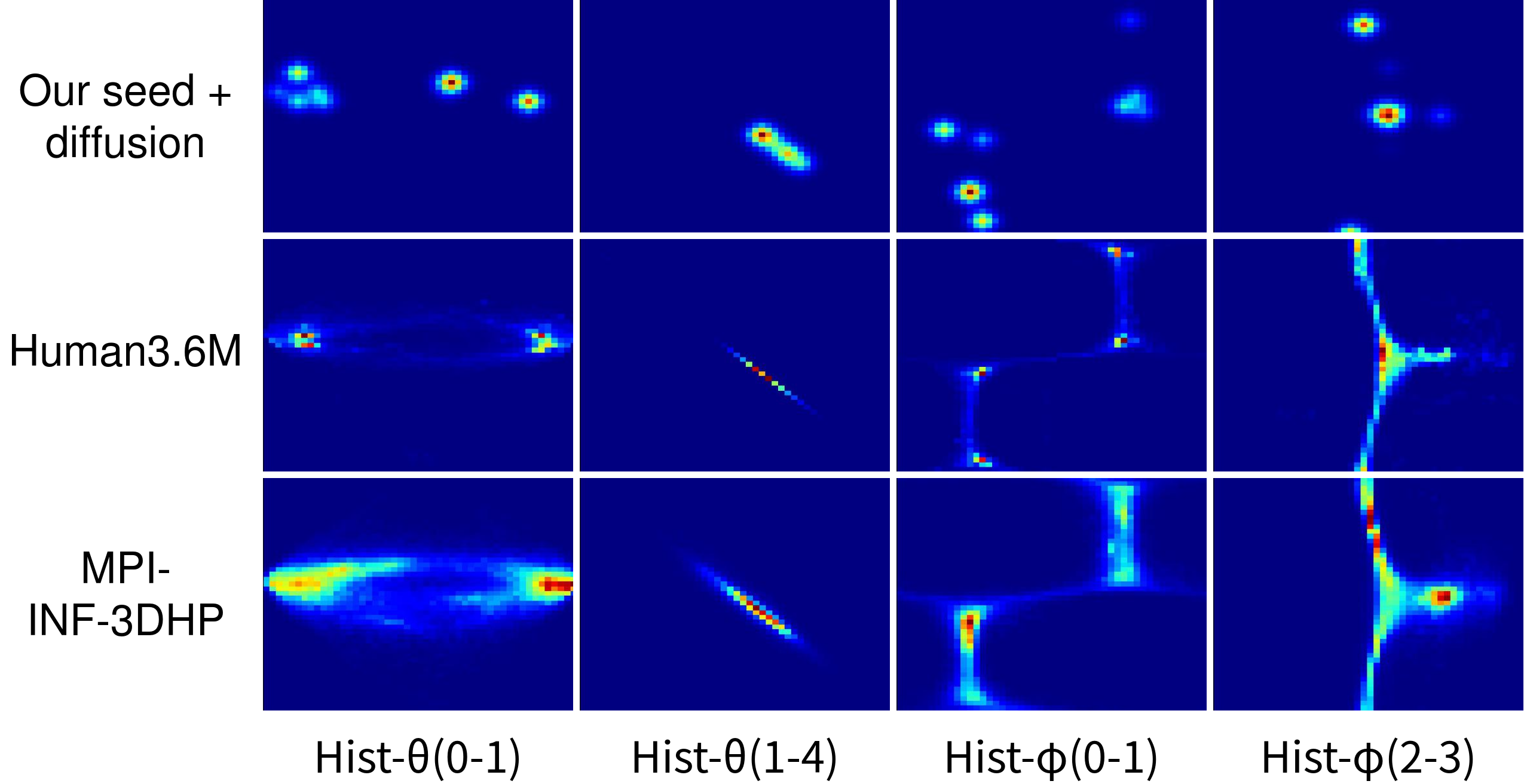}
  \includegraphics[width=0.45\linewidth]{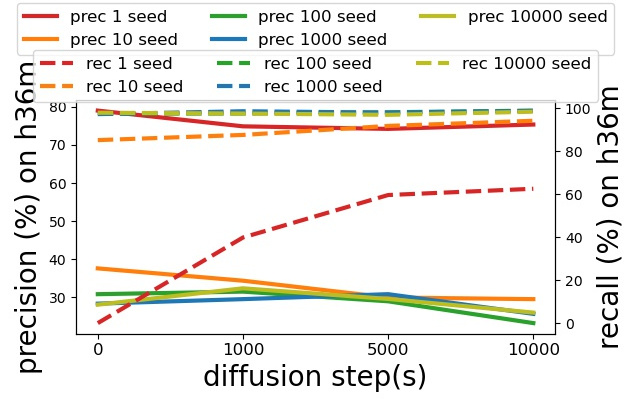}
  \caption{\textbf{Left}: Examples of distributions of angle $\theta$ and $\phi$ from same parent-child pairs computed on Human3.6M, MPI-INF-3DHP, and our diffusion process. \textbf{Right}: Precision and recall evaluated with 5k generated samples and 5k real 2D samples from h36m.}
  \label{fig:h36m-3dhp-distribution}
\end{figure}

We want to see how similar our synthetic skeletons are to real skeletons. Qualitatively we compare our distribution after diffusion with the distribution of the whole Human3.6M and MPI-INF-3DHP datasets, for some of the joints as shown in \autoref{fig:h36m-3dhp-distribution}. We can see that, even though there are many poses in MPI-INF-3DHP have never appear in Human3.6M, the distributions of angles $\theta$ and $\phi$ of these two real datasets have very similar shapes, which means our local spherical coordinate system successfully models the invariance of the biological achievable human pose angles and their frequencies which are independent of camera view point.
Our seeds+diffuse strategy produces a Gaussian mixture which succeed in covering big parts of real dataset's distribution.

Quantitatively we apply a precision/recall test, as is common practice with GANs\cite{ferjad2020icml(prdc)}. We sample 5000 real and 5000 synthetic poses and project them to 2D plane using the scaleless projection in \ref{eq:2Dproj} and the Euclidean distance. Precision (resp. Recall) is defined as percentage of synthetic samples (resp. real samples) inside the union of the balls centered on each real sample (resp. synthetic sample) and with a radius of the distance to its 10-th nearest real sample neighbor (resp. synthetic sample neighbor). In our case, we already know that most synthetic skeleton generated by our Markov tree are biologically possible thanks to the limits in the generation intervals. As such, we are more interested in a very high recall so as to not miss the diversity of real skeletons. All our seed sets have more than $70\%$ recall and highest one achieves $91.8\%$ recall. The precision, on the other hand, is around $40\%$, with $47.1\%$ as the highest, which is still good considering we only start with 10 manually lifted initial poses for each seed.

\subsection{Effect of diffusion}

We want to see why diffusion process is essential to our method. We take respectively 1, 10, 100, 1000 and 10000 samples of 3D poses
on Human3.6M dataset as initial seed to make distribution graphs, and apply our 2D precision recall test after diffusion process. The result is shown in \autoref{fig:h36m-3dhp-distribution}. We can see that diffusion generally increase recall value at the cost of precision value. The distribution using 1 samples as seed is much worse with the others in recall which means it can only cover around $60\%$ of samples from real dataset even with diffusion process, while the distribution using 100 samples or more are close in performances. The diffusion process can reduce the gap between the distribution using 10 samples as seeds and those using 100 or more samples, which is important to us considering we want to avoid handcrafting a lot of initial poses.

\subsection{Layout adaptation}

We show that our synthetic generation and training method also work on a different keypoint layout by applying the whole process on a newly defined hierarchic Markov tree based on 24 keypoints of SMPL model \cite{SMPL:2015(SMPLModel)} and evaluating on 3DPW dataset\cite{Marcard_2018_ECCV(3DPW)}. We use 24 samples from its training set (one frame from each video) using our 2D variance based criterion for the seeds. Since our training method is scaleless, we rescale the predicted 3D poses by the average Forbenius norm of the 24 samples in the seed. The average MPJPE of 10 different seeds is shown in \autoref{tab:abla_3DPW}. This validates the generalization capability of our method.

\begin{table}[t]
    \small
    \centering
    \begin{tabular}{l|c| c }
    Method & Labeled training data & MPJPE$\downarrow$\\
    \hline
    CLIFF (ECCV22) & H36m + 3DHP + COCO + MPII + 3DPW & \textbf{52.8}\\
    DynaBOA (TPAMI22) & H36m + 3DPW & 65.5\\
    \hline
    ours & 24 samples from 3DPW & \underline{61.09 $\pm$ 2.16} \\
    \end{tabular}
    \caption{Results on the 24-keypoint SMPL model, compared to the state-of-the-art}
    \label{tab:abla_3DPW}
\end{table}

\section{Conclusion}

We present an algorithm which allows to generate synthetic 3D human skeletons on the fly during the training, following a Markov-tree type distribution which evolve through out time to create unseen poses.
We propose a scaleless multiview training process based on purely synthetic data generated from a few handcrafted poses. We evaluate our approach on the two benchmark datasets Human3.6M and MPI-INF-3DHP and achieve promising results in a zero shot setup.

\clearpage

%
%
%
\bibliographystyle{splncs04}
\bibliography{egbib}
\end{document}



\begin{figure*}[t]
  \centering
  \includegraphics[width=0.9\linewidth]{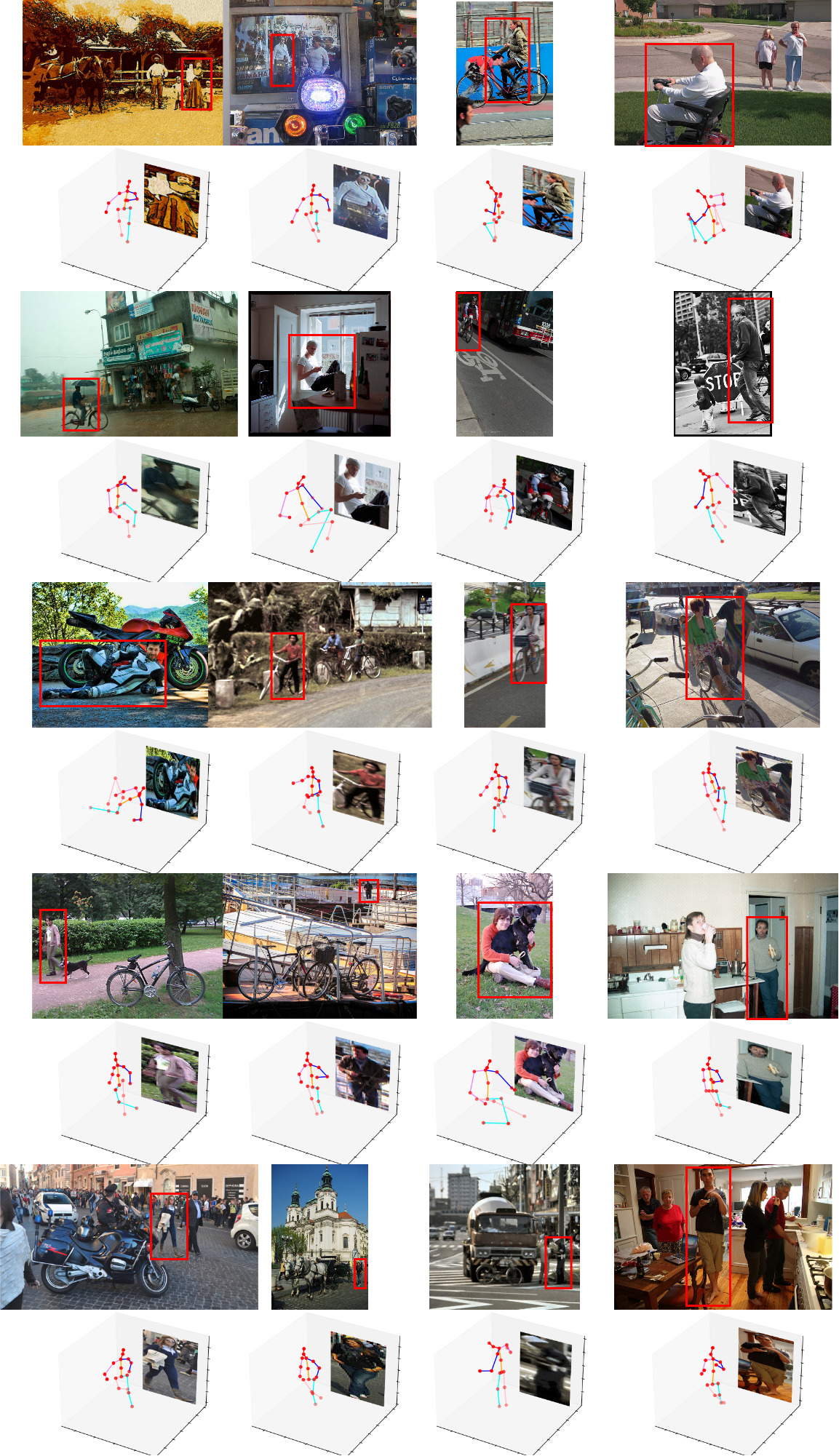}
  \caption{Our model inferenced on images from COCO dataset. The head position is not always accurate, because the input skeleton is an interpolation from COCO skeleton to Human3.6m skeleton, which miss the top keypoint of the head.}
\end{figure*}
\begin{figure*}[t]
  \centering
  \includegraphics[width=0.9\linewidth]{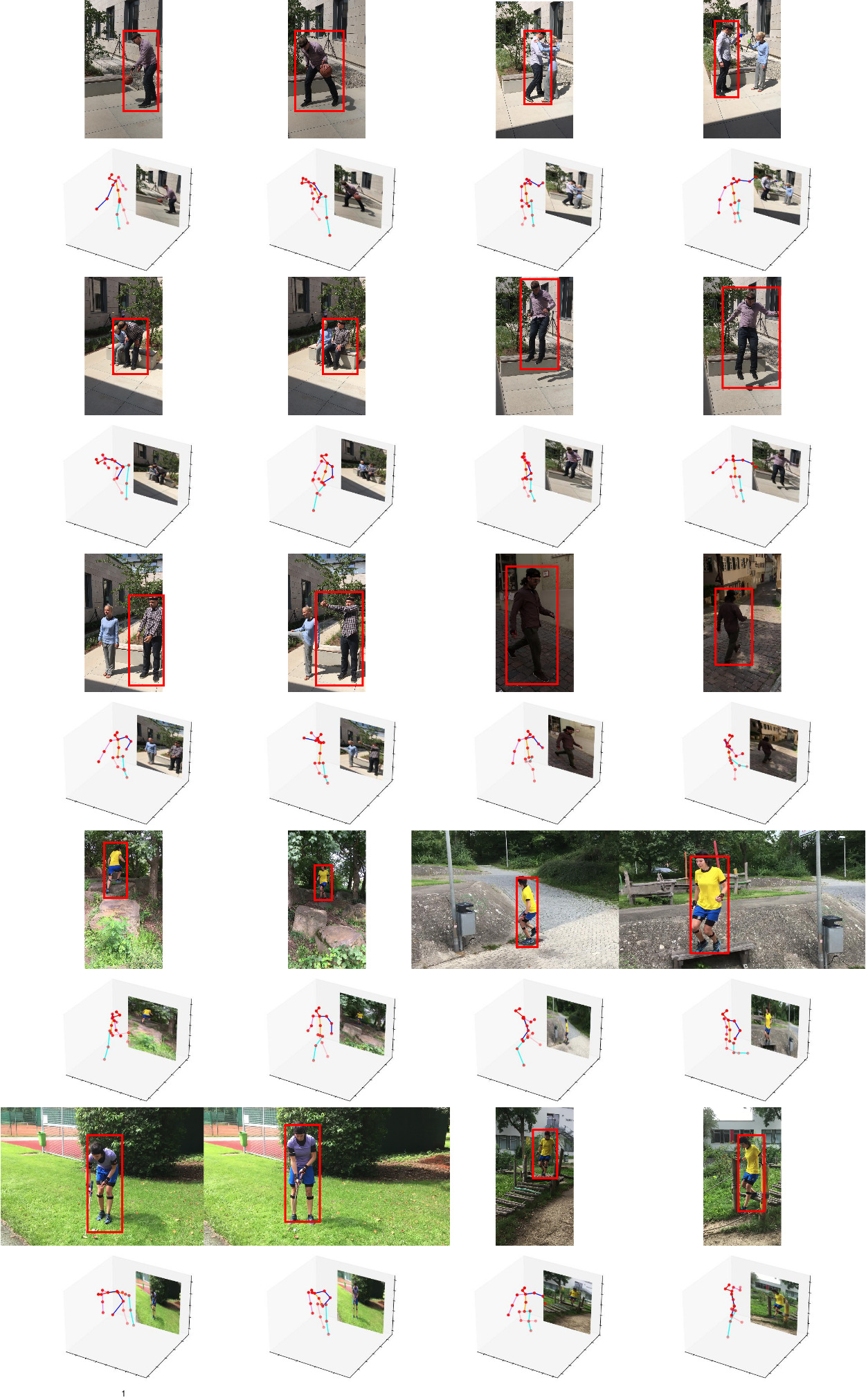}
  \caption{Our model inferenced on images from 3DPW validation dataset. One problem we have when doing inference on 3DPW is that our model is trained with 2D poses with all keypoints given, while 3DPW dataset has many keypoints not annotated even in 2D, in this case the lifted pose doesn't perform well.}
\end{figure*}